# AI-Based Framework for Understanding Car-Following Behaviors of Drivers in A Naturalistic Driving Environment

Armstrong Aboah[1, *], Abdul Rashid Mussah[2], Yaw Adu-Gyamfi[3]

[1]Civil and Environmental Engineering Department, University of Missouri-Columbia, USA,

E25O9 Lafferre Hall, Columbia, MO 65211

*Corresponding Author's Email: aa5mv@umsystem.edu

[2]Civil and Environmental Engineering Department, University of Missouri-Columbia, USA,

E25O9 Lafferre Hall, Columbia, MO 65211. Email: aakmx@umsystem.edu

[3]Civil and Environmental Engineering Department, University of Missouri-Columbia, USA,

E25O9 Lafferre Hall, Columbia, MO 65211. Email: adugyamfiym@missouri.edu

Abstract

The most common type of accident on the road is a rear-end crash. These crashes have a significant negative impact on traffic flow and are frequently fatal. To gain a more practical understanding of these scenarios, it is necessary to accurately model car-following behaviors that result in rear-end crashes. Numerous studies have been carried out to model drivers' car-following behaviors; however, the majority of these studies have relied on simulated data, which may not accurately represent real-world incidents. Furthermore, most studies are restricted to modeling the ego-vehicle's acceleration, which is insufficient to explain the behavior of the ego-vehicle. As a result, the current study attempts to address these issues by developing an artificial intelligence (AI) framework for extracting features relevant to understanding driver behavior in a naturalistic environment. Furthermore, the study modeled the acceleration of both the ego-vehicle and the leading vehicle using extracted information from NDS videos. According to the study's findings, young people are more likely to be aggressive drivers than elderly people. In addition, when modeling the ego-vehicle's acceleration, it was discovered that the relative velocity between the ego-vehicle and the leading vehicle was more important than the distance between the two vehicles (car-following distance).

## 1. Introduction

The rear-end crash is the most common type of accident encountered on the highway, according to statistical data for related accidents. These crashes have a significant negative impact on the flow of traffic and typically result in serious repercussions (Zheng and Sarvi, 2016). An examination of the sequence of events that lead up to rear-end crashes can be broken down into two distinct scenarios. The first scenario involves situations whereby there is minimal headway between vehicles, creating an environment whereby a sudden braking event by the leading vehicle results in a potential rear-end collision by the following vehicle, even if the brakes of the following vehicle are applied on time. This car-following phenomenon is described as tailgating. In the second scenario, the driver is maintaining a car-following distance that is relatively safe; however, impairments caused by distractions or fatigue results in the driver of the following vehicle not being able to apply their brakes in time. This scenario is similar to instances whereby there is absolutely no braking action taking place by the following vehicle. Other edge-case situations can be described in the context of anomalous driving behaviors, such as speeding and aggressive acceleration events. To gain a better understanding of the aforementioned scenarios from a practical standpoint, it is necessary to accurately model car-following behaviors that lead to rear-end crashes, and that is what this study seeks to achieve.

The aforementioned scenarios of rear-end crashes each have a significantly high incidence rate. As a result, some automobile companies have begun encouraging the installation of a warning system for rear-end crashes. Alarm systems for rear-end crashes are able to monitor the motion of the car in front of them by using radars or visual sensors, and they can sound an alarm if they detect an impending risk. Some drivers may develop a hostile attitude toward other drivers during car following if the alarms are set off too frequently, given that every driver has a different approach to driving (Hoogendoorn et al., 2010). In order to reduce the likelihood of the driver becoming distracted by the alarm, or desensitized to its warning chimes, these systems typically set the timer for the alarm to go off when the vehicle is already in a relatively hazardous state. However, the timing of the warnings has a significant impact on how well the warning system works. There is a potential for an increase in the number of accidents if the warning time is delayed (Tang and Yip, 2010). This discussed challenge could easily be resolved if the acceleration of the leading vehicle can be correctly estimated relative to the acceleration of the ego-vehicle. The current study thereby explores the potential of developing models utilizing naturalistic driving video data that are capable of predicting the acceleration of the leading vehicle as well as the ego-vehicle.

*1.1 Motivation*

Although a number of studies have been conducted to model the car-following behaviors of drivers, the majority of these studies have relied on simulated data that may not accurately represent incidents that occur in the real world. In addition, very few longitudinal studies have been conducted on the car-following behavior of drivers in naturalistic environments. These studies, however, are limited to developing models that can only estimate the acceleration of the ego-vehicle, which is insufficient to explain the behavior of the ego-driver. This limitation exists because data related to the acceleration of the leading vehicle is rarely available when using a naturalistic driving study (NDS) dataset. As such, the current study attempts to address this issue by developing an AI framework capable of processing and extracting parameters from NDS videos necessary to model the driving behavior (acceleration) of the leading vehicle as well as the ego-vehicle. In addition, there have been no previous longitudinal studies carried out on the car-following behavior of various driver demographics in a naturalistic environment. Given the capacity of this framework, the potential to accomplish this feat opens up to the world of safety practitioners in their continued assessment of driver behavior. Such research is important because it enhances our understanding of the driving styles of various demographic groups and the causes of crashes, beyond what can be gleaned from crash data. The study addresses this deficiency by conducting longitudinal studies of different demographic groups of drivers.

*1.2 Contributions*

In light of the gaps identified in previous studies, the study seeks to primarily develop an AI framework to extract parameters from naturalistic driving videos that provide better insight into drivers' behavior in a naturalistic environment. To achieve this goal, we formulated three objectives described below.
1. First, the study develops a framework for extracting features pertinent to comprehending the behavior of drivers in natural environments. To achieve this objective, we investigated a number of video depth estimation techniques for determining the distance between the leading vehicle and the ego-vehicle. Monocular depth estimation models were validated and calibrated utilizing ground-truthed lidar depth data. Moreover, we determine the relative velocities of the leading vehicle and the ego-vehicle using optical flow. The study further relied on the relative velocity, car-following distance and the acceleration of the ego-vehicle to estimate the acceleration of the leading vehicle.
2. Second, we demonstrated how this framework can allow us to analyze the car-following behaviors of various demographic groups. To accomplish this objective, numerous visualization plots and statistical tests were conducted. The study employed the sample mean t-statistics to determine whether different demographic groups exhibit distinct driving behavior.
3. Third, we modeled the acceleration of both the ego-vehicle and the leading vehicle using a machine-learning algorithm. We utilized the XGBoost algorithm to develop the various acceleration models. The explanatory variables used in the models were car-following distance, relative velocity, and ego-vehicle acceleration (only used to model the acceleration of the leading vehicle). We investigate further the variables that best explain the leading and ego-vehicle accelerations.

*1.3 Outline*

The rest of the paper is organized as follows. Section two provides a review of relevant literature. The data used for this study is presented in Section three. Section four presents the methodology employed by this study. Section five presents a discussion of the results of the model development. Finally, Section six presents a summary of the research, the conclusions drawn from the results, and recommendations for future research.

## 2. Related works

Understanding driver car-following behaviors is a critical step in developing crash countermeasures to reduce rear-end collisions. Over the last two decades, there has been an enormous amount of research into understanding and modeling driver car-following behavior. As a result, this section provides reviews of car-following studies as well as other relevant literature pertaining to the current research work. First, we review various studies on car-following behavior conducted in the past two decades. Second, we go over the various video processing techniques such as monocular depth estimation approaches that have been used in previous studies.

*2.1 Modeling Car-following Behaviors*

On the topic of car-following safety, a great number of studies have been conducted, each of which has approached the topic from a slightly different direction. Kometani, (1959) came up with the idea for a car-following model that was based on safety distance. This distance is the shortest one that must be traveled in order to avoid a crash in the event that the vehicle in front of you begins to use its emergency brakes. Car-following was broken up into two phases according to Treitere et al., (1974) who began their research with the General Motors model as their foundation. This process involves both quickening and slowing down the speed at various points. Car-following, according to Aron, (1988), can be broken down further into 3 distinct stages: deceleration, acceleration, and maintaining. Helly, (1959) recommended making use of a linear model that takes into account the distance traveled in addition to the acceleration and the relative speed. Peter, (1998) made a suggestion for a model for the desired spacing based on the findings of the research concerned with desired spacing distance. Numerous studies concentrated their attention on the car-following model with regard to the speed of the vehicles and the distance between them. Following the experiment, Jiang et. al. (2015) carried out a motorcade in order to investigate the relationship between the speed of a car and the distance it kept behind another car in a number of different types of traffic. When following a vehicle, it is extremely important, as stated by Tang et al.(2017), to take into consideration both the forward and the backward safe distances. The findings of the study indicate that models which took into account both types of safe distances performed appreciably better than those which did not (Tang et. al., 2017). Gipps, (1981) laid the groundwork for subsequent research on car following by providing a theoretical framework in which he proposed a driver model that could be simulated, and which followed a car. A correlation analysis was carried out by making use of parameters derived from the actual flow of traffic (Gipps, 1981). Because of this, future research on car following will have a theoretical foundation to build upon. Zhou et al. developed a more reliable model for car following by taking into account the motion characteristics of the vehicle in front of them (Zhou et. al., 2014). According to the findings of Tang et al. (2014), Yang et al. (2017), and Tordeux et al. (2010), the headway is an important parameter that plays a role in the evaluation of the driver's level of risk. In addition to the rate of driving, another factor to take into account is the time headway (THW)(Aboah and Adu-Gyamfi (2020); Aboah A. (2019); Shoman et al. (2020); Aboah et al. (2022); Osei et al. (2019); Aboah and Arthur (2021)).

*2.2 Monocular Depth Estimation*

Over the recent years, there have been many deep learning networks and their variants that perform monocular depth estimation (Eigen et al. (2014); Zheng et al. (2018); Yin et al. (2019); Ranftl et al. (2020); Wang et al. (2020)). These networks provide superior performance with various architectural transformations, data augmentation strategies, and innovative cost functions. Many of these neural networks however do not focus on providing high-resolution/quality dense depth maps (Silberman et al. (2012); Wadhwa et al. (2018); Wang et al. (2018)). Some of them which produce high-quality depth maps do require complex architectures. These complex networks use numerous deep layers, residual modules, guiding modules, sequential modules, attention-based modules, etc. (Laina et al. (2016); Xu et al. (2017);

Xu et al. (2018); Fu et al. (2018); Swami et al. (2020)) to achieve the complex task of monocular depth estimation. Laina et al., (2016) use a deep layer CNN confining residual learning to improve output depth map resolution. Xu et al., (2017) propose a novel sequential framework that fuses multi-scale convolutional neural network with continuous conditional random field module for accurate depth maps. Hao et al., (2018) model use dilated convolutions and attention mechanism for extracting multi-scale feature from input image while maintaining dense feature maps and fuse multi-scale features respectively to produce high-quality depth map. Xu et al., (2018) use a multi-scale convolutional neural network along with a conditional random field and structural attention module. Fu et al., (2018) propose a deep ordinal regression network that discretizes depth and remodels learning of the depth estimation network as an ordinal regression problem. Swami et al., (2020) is an improved version of the Fu et al., (2018) which use a fully differentiable ordinal and pixel-wise regression network along with a depth refinement module for improved depth estimation. Recently, there has also been a lot of research to achieve high-quality monocular depth estimation using Transformer architecture (Xie et al. (2020); Ranftl et al. (2021)) due to its success achieved in other computer vision-related tasks like image classification, image segmentation, etc. Also, the same was observed with deep encoder-decoder (Ummenhofer et al. (2017); Zhou et al. (2018)) type structures. However, these approaches are computationally expensive which impedes their embedding in edge technology. Therefore, we aim to analyze simple architectures with less computation and comparable state-of-the-art performance for the task of monocular depth estimation (Ibraheem et al. (2018); Doyeon et al. (2022); Miangoleh et al. (2021)).

### 3. Data

The data collection for this study was carried out with the assistance of Blackbox sensors, which were created by Digital Artifacts LLC. The sensors were put into individual, privately owned vehicles with the purpose of continuously recording activities that took place both inside and outside of the vehicle. The sensor instrumentation incorporates a number of sensors, including high-resolution cameras, infrared sensors, high-precision GPS, and wireless onboard diagnostics (OBD). The sensor package that is mounted on the windshield can be seen in Figure 1. This package is mounted behind the rear-view mirror. The system is equipped with two cameras, one of which captures 1) a view of the roadway in front of the vehicle at all times, and 2) a view of the driver as well as the interior of the vehicle. From the moment the ignition key is turned to the moment it is turned off, the behavior of the driver is being continuously recorded. The research involved 77 participants and was conducted over the course of three months. Data was gathered on a total distance of 289,682 kilometers in multiple regions of the United States of America. This dataset contains much more detailed information on driver behavior across a wide range of geographic environments than laboratory-based or retrospective studies are able to provide.

This study utilized data from a handful of individuals (two elderly drivers and two young drivers) constituting a total of 65 minutes of driving video data over a combined journey of 56 kilometers, and with processing carried out on over 38000 video frames.

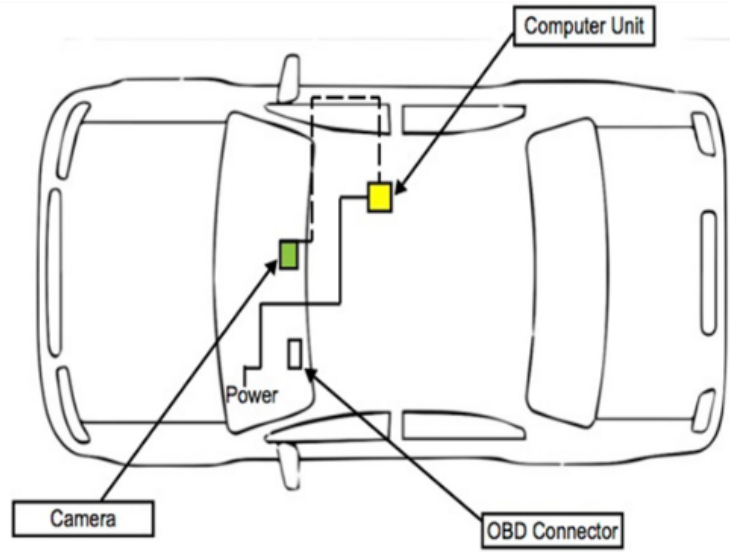

Figure 1: The Positioning of the Blackbox sensors in the vehicle

## 4. Methodology

The general methodological framework shown in Figure 2 can be grouped into two main stages. First, we extract the car following distance using distance obtained from the monocular depth estimation and tracking of the leading vehicle (LV). Second, we estimate the acceleration of the LV by combining the estimated car-following distance, relative velocity and the acceleration of the ego-vehicle.

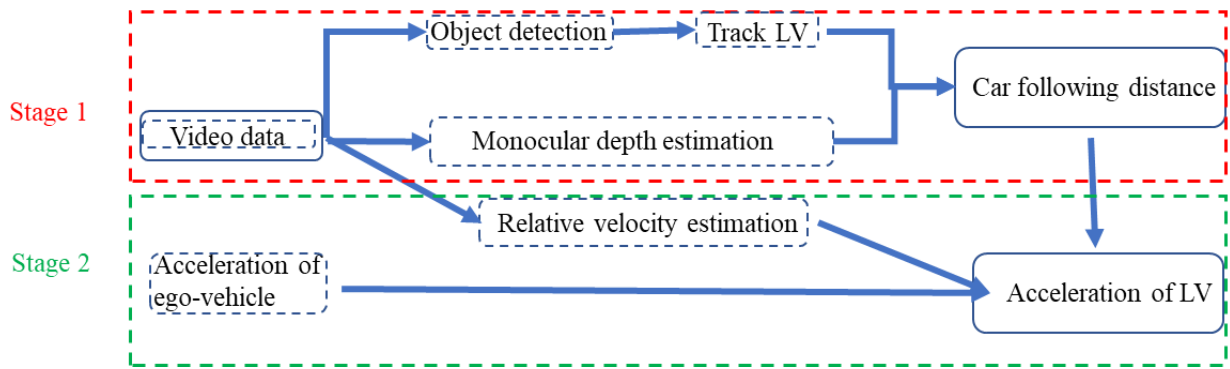

Figure 2: Methodological Framework

### 4.1. Extraction of car-following distances

The process involved in the extraction of car-following distances can be divided into three major steps. First, train an object detection model to detect vehicles and traffic signs. Second, using a trained monocular depth estimation model to estimate the scene depth of the video images. Finally, combine the first two steps with a heuristic algorithm for tracking leading vehicles to estimate the car-following distance.

*4.1.1 Object detection*

The state-of-the-art single-stage object detection algorithm used for developing the detection model is YOLOv5 (Aboah et al. (2021); Shoman et al. (2022)). The YOLOv5 network consists of three main pieces viz. Backbone, Neck and Head. The Backbone consists of a convolutional neural network that bundles and forms image-representational features at contrasting granularities. The architecture's neck consists of a series of layers that blends and integrates image representational features to proceed further with prediction. Similarly, the head utilizes features from the neck and gets hold of box and class prediction functionality. CSPDarknet53 backbone within YOLOv5 contains 29 convolutional layers $3 \times 3$, receptive field size of $725 \times 725$ and altogether 27.6 M parameters. Besides, the SPP block attached to YOLO's CSPDarknet53 expands the proportion of receptive fields without influencing its operating speed. Likewise, the feature aggregation is performed through PANet by exploiting different levels of backbone. YOLOv5 pushes state-of-the-art by using features such as the weighted-residual-connections, cross-stage partial-connections, cross mini-batch, normalization and self-adversarial training, making it exceptionally efficient. In the current study, we trained and deployed our YOLOv5 model on the PyTorch framework. To further accomplish the task of vehicle detection, the YOLOv5 model is fine-tuned by adjusting to the following hyperparameters: batch-size 64, the optimizer weight decay value of 0.0005, setting the initial learning rate of 0.01 and keeping the momentum at 0.937.

*4.1.2 Monocular depth estimation*

The study utilized different monocular depth estimation models developed by [1,2,3] to estimate the car's following distance. Each model is explained in detail in the sections below

a) High quality monocular depth estimation via transfer learning: A simple encoder-decoder UNet type of machine learning model was used for the task of high-quality depth estimation. Fig. 3. shows the encoder-decoder neural network model [1]. The encoder used in the model is the truncated encoder structure of high-performing DenseNet169 architecture. For this encoder, the DenseNet169 model is truncated with the top layers leaving the bottom deep layers. The DenseNet169 is pretrained on the ImageNet dataset for the object classification task. This process of leveraging a high-performance neural network trained on different tasks for initializing the model with the more complex tasks is called transfer learning.

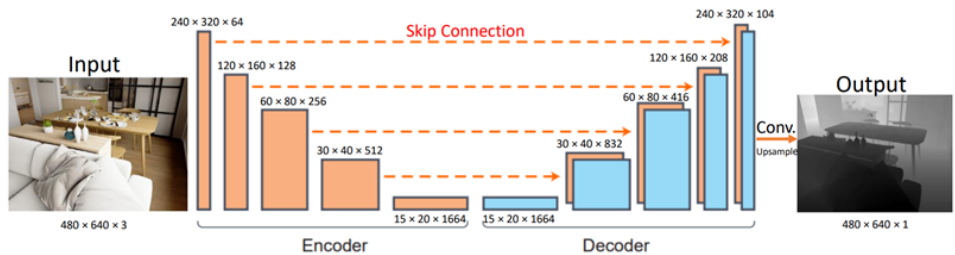

Figure 3: Pipeline of the Proposed Method

The encoder consists of the 3x3 convolution layers and downsampling layers which decrease the feature size and increase the number of feature maps along the process to get the final bottleneck features of the image. The encoder also consists of the dense and transition block which is one of the advantages of the DenseNet169 networks. The dense block performs convolution and pooling operation where each processing node gets the input from all the previous node outputs. The dense block, therefore, encourages feature propagation and reuse, which can be useful for understanding. The dense block reduces the feature size and increases the number of feature maps which can increase the trainable parameters. Therefore, to take care of this, the transition block reduces the number of feature maps of the dense block output. Once, we get the bottleneck features from the encoder. These features are given as input to the decoder.

The decoder consists of the up-sampling layers, which use bilinear sampling, and convolution layers. The convolution layers get the feature input from the previous layer of the decoder output and from the same scale encoder convolution layer output. At each layer the decoder performs up-sampling and convolution which decreases the number of feature maps to half and increases the feature map size. The model also consists of skip connections which concatenate the encoder features to the decoder. The skip connections help to regain the details that can be lost during down sampling. It also helps to avoid vanishing and exploding gradient issue as the gradient can travel directly from decoder to the corresponding scale encoder using skip connection during backpropagation. Finally, the decoder predicts the depth map at the output.

$$L(y, \hat{y}) = \lambda L_{depth}(y, \hat{y}) + L_{grad}(y, \hat{y}) + L_{SSIM}(y, \hat{y}) \quad (1)$$

$$L_{depth}(y, \hat{y}) = \frac{1}{n}\sum_{p}^{n} |y_p - \hat{y}_p| \quad (2)$$

$$L_{grad}(y, \hat{y}) = \frac{1}{n}\sum_{p}^{n} |g_x(y_p, \hat{y}_p)| + |g_y(y_p, \hat{y}_p)| \quad (3)$$

$$L_{SSIM}(y, \hat{y}) = 1 - \frac{SSIM(y, \hat{y})}{2} \quad (4)$$

The loss function used for the training has a significant effect on the overall performance of the model. For this model, three terms in the loss function are used as shown in equation (1). The first term $L_{depth}$ shown in equation (2) defines the general depth reconstruction error that tries to minimize the difference between the ground truth and the predicted depth map. This is the common loss term used by the depth estimation models, where n is the total number of pixels in depth map, is the ground truth pixel depth value, is the predicted pixel depth value, and λ weight is taken equal to 0.1. The second loss term $L_{grad}$ shown in equation (3) penalizes the model if there are distortions of high-frequency details like boundaries of the objects in the scene. This is done by calculating depth map gradient g() in x and y directions. The loss term then will ensure that the depth map gradient of the ground truth and prediction are as close as possible. The third loss term $L_{SSIM}$ shown in equation (4) uses SSIM function to calculate the similarity between the ground truth and the predicted depth map. 1-SSIM means the dissimilarity value which ensures that the dissimilarity between the ground truth and the predicted depth map is as low as possible.

b) Global-Local Path Networks for Monocular Depth Estimation with Vertical CutDepth: The study used an encoder-decoder model shown in FigureThe model uses a transformer encoder and decoder with skip connections called selective feature fusion (SFF) module to capture the global image context and local connectivity respectively. The proposed model is a global-local path network that extracts significant features on diverse scales and effectively delivers them throughout the network.

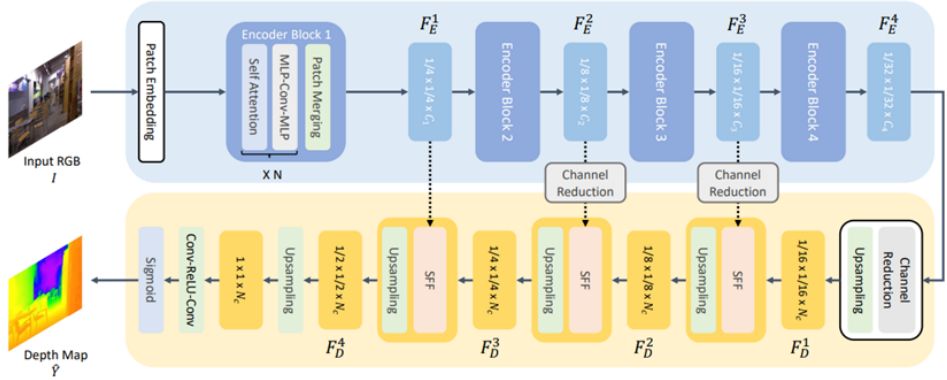

Figure 4: Pipeline of the Proposed Method

The hierarchical transformer is used as the encoder to capture global relationships. First, the input image is embedded as a sequence of patches with a 3 × 3 convolution operation. These embedded patches are then used as an input to the transformer block, which consists of multiple sets of self-attention and the MLP-Conv-MLP layer with a residual skip. Finally, the transformer output undergoes patch merging with overlapped convolution. The model uses four transformer blocks to get multi-scale features.

The decoder with an effective fusion module is then used to capture local features to produce a fine depth map while preserving structural details. For this, the bottleneck features channel dimensions from the encoder are first reduced using 1 x 1 convolution. Then the decoder uses consecutive bilinear up-sampling to enlarge the feature size and SFF module. Finally, the output is passed through two convolution layers and a sigmoid function to predict the depth map.

To exploit the local structures having fine details, the network combines the encoded and decoded features using skip connections with the input-dependent feature fusion module SSF. The SFF module helps the network to selectively focus on the salient regions by estimating the attention map for both encoder and decoder features. For this first, the channel dimensions of the decoded features FD and encoded features FE are matched with the convolution layer and provided as input to the SFF module. The SFF then concatenates these features along the channel dimension and pass it through two 3 × 3 Conv-batch_normalization-ReLU layers. Finally, the convolution and sigmoid layers produce a two-channel attention map which are then multiplied to each local and global feature to focus on the significant locations. These multiplied features are then added elementwise to construct a hybrid feature.

$$L = \frac{1}{n} \sum_i d_i^2 - \frac{1}{2n^2} \sum_i d_i^2 \quad (5)$$

$$d_i = \log \log y_i - \log \log y_i^* \quad (6)$$

The loss function for the model uses scale-invariant log scale loss as shown in equation (5) and (6) where and indicates the $i^{th}$ pixel value in the ground truth and predicted depth map.

C) Boosting monocular depth estimation models to high-resolution via content-adaptive multi-resolution merging: The architecture of the model is such that a double-depth-estimation network is analyzed that combines the two depth estimations of the same image at different resolutions adaptive to the image content to generate result with high-frequency details while maintaining the structural consistency. It is observed that the low-resolution input to the network produces structural consistent depth maps as they learn the overall global content in the image while the high-resolution input captures the high-frequency details but loses the overall structure of the scene generating low-frequency artifacts in the depth estimate. The proposed model therefore embeds the high-frequency depth details of the high-resolution patches into the structural consistent depth of the small resolution input that provides a fixed range of depths for the full image.

First the high-resolution inputs are created by selecting the patches from the input image of resolutions adaptive to the local depth cue density to be fed to the network. For creating image adaptive patch resolution containing contextual cues, image edge map is computed, using gradients and thresholding, as edge maps are correlated to the contextual hues. The edge map is then used to determine the maximum resolution where every pixel in the patch has contextual information. Finally, the patches are selected by initiating their size equal to the receptive field. Their size is then increased if the edge density is higher than the image density until their density becomes equal while the patches with less edge density are discarded.

The created high-resolution patches and the low-resolution image are now provided as input to the network to produce depth estimates. Then the depth estimate of the patches is combined into a low-resolution structurally consistent base depth estimate to achieve a highly detailed high-resolution depth estimation. For this, the model uses a standard Pix2Pix network architecture with a 10-layer U-Net. The 10-layer U-Net aims to increase the training and inference resolution, as this merging network will be used for a wide range of input resolutions. The network is finally trained to transfer the fine-grained depth details from the high-resolution input patches to the low-resolution input depth estimate.

*4.1.3 Determining the leading vehicle*

We relied on a set of heuristic algorithms to determine the leading vehicle among all vehicles identified in an image. To begin, we determined that the image's center was roughly the vanishing point of the road lane. Multiple iterations revealed that multiplying the image's base by a factor of 0.2 and 0.8 yields a triangulation that corresponds to a definition of the perspective of the lane of the ego-vehicle. We compute the gradient for each base coordinate point and the center point (vanishing) of the line. Then, we determine if the predicted vehicle bounding boxes' base coordinates lie within the triangle. If both base points fall within the triangle, the vehicle is identified as the leading vehicle.

*4.2 Estimating the Acceleration of the Leading Vehicle*

In order to estimate the acceleration of the leading vehicle, we first estimated the relative velocities between the leading vehicle and the ego-vehicle using optical flow. We then adopted Newton's second equation of motion which relates distance, velocity, and acceleration as shown in Equation 1. We modified Equation 7 to reflect what is seen in Equation 8. Where the change in distance is the car following distance between the ego-vehicle and the leading vehicle. Whereas the relative velocity between the ego-vehicle and the LV was used as the change in velocity in that equation. Finally, the change in acceleration was modeled as the difference between the acceleration of the ego-vehicle minus the acceleration of the LV. From Equation 8, we are able to estimate the acceleration of the leading vehicle.

$$s = ut + \frac{1}{2}at^2 \quad (7)$$

$$\Delta s = \Delta ut + \frac{1}{2}\Delta at^2 \quad (8)$$

## 5. Results

*5.1 Comparative analysis of monocular depth estimation methods*

In this study, three monocular depth estimation models were evaluated and compared as shown in Figure 5. The depth estimation of each model was compared to the true depth as determined by a lidar scan in order to validate the accuracy of the model for the task at hand. We calculate the root mean squared error depth between the ground truth (lidar data) and each model's estimated depth. The model with the least root mean squared error was selected as the optimal model for the study.

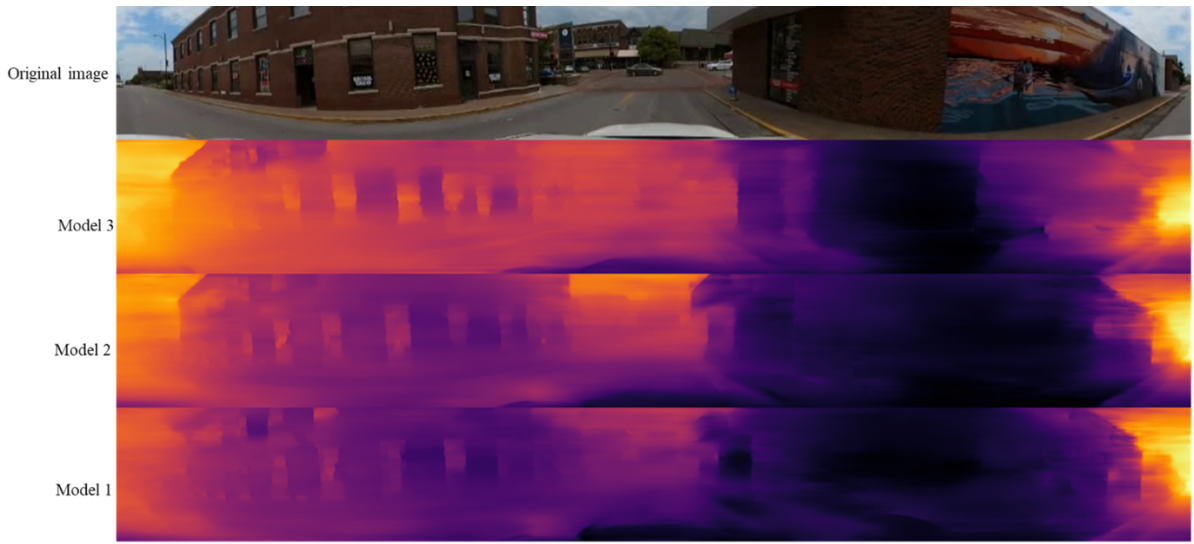

Figure 5: Comparing depth estimation of the three monocular depth estimation models

Figure 6 compares the estimated distances to the actual distances for each model. Model 1 performs better at predicting the depth of objects when they are closer, but utterly fails when they are farther away. The same could be said for Model 2, with the exception that it performs marginally better than Model 1 when predicting the depth of distant objects, as illustrated in Figure 6. Model 3 has the greatest correlation between predicted and actual distances when compared to Models 1 and 2. The study further analyzes these models by comparing the root mean square error.

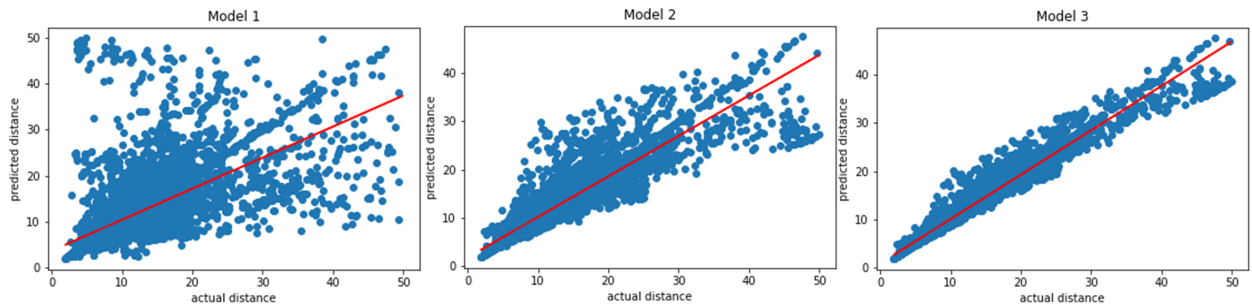

Figure 6: Comparative analysis of the prediction accuracy of

Table 1. compares the root mean squared error of the three models. Model 3 had the least squared error of 1.79. Followed by model 2 of root mean squared error of 58 and lastly model 1 of root mean squared error of 6.23. Based on the results from Table 1. The study settles on model 3 for the rest of its analysis.

Table 4. 1: Root mean squared error of three monocular depth estimation

| model | Model 1 | Model 2 | Model 3 |
|---|---|---|---|
| Root-mean-squared-error | 6.23 | 58 | 1.79 |

*5.2 Estimation of car following distance*

The estimation of the car-following distance was a multi-step approach as described in the methodology. First, we estimated the scene depth using monocular depth estimation (Model 3). Figure .7 below shows the input image and output image after the input image was passed through the monocular depth estimation model to estimate the scene depth.

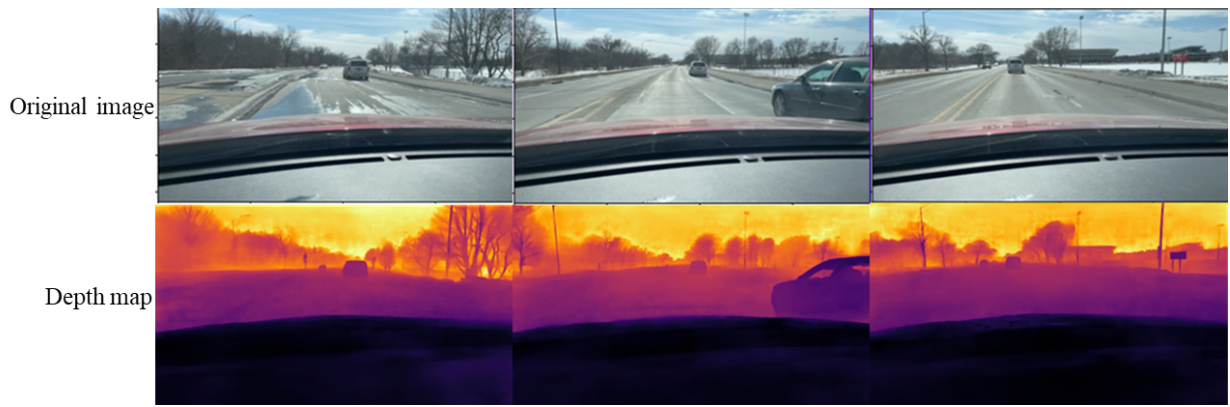

Figure 7: Input image (row 1) and output image (row 2) after the input image was passed through the monocular depth estimation model to estimate the depth map

Second, we detected relevant objects in the images by using a single-stage object detection model YOLOv5. The results of the detection stage are shown in Figure .8 below. The objects identified in this stage were cars, trucks, traffic signs and traffic signals.

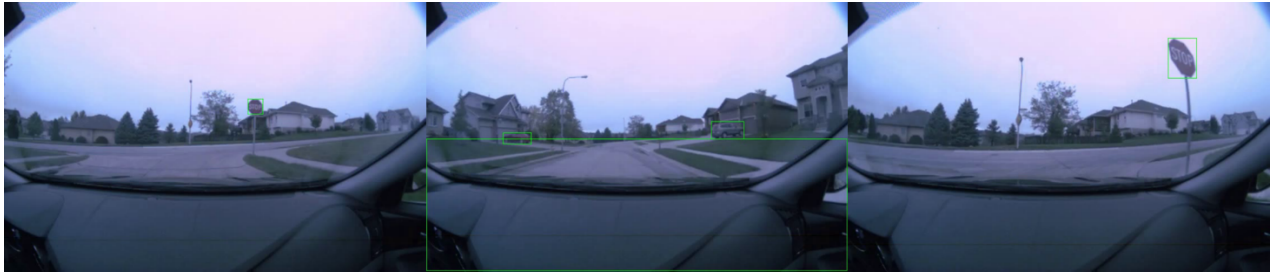

Figure 8: Object detection using YOLOv5

Finally, we combined the first and second results with a heuristics algorithm to determine if a detected car or truck is in front of the ego-vehicle or not. The results of that are shown in Figure .9 below.

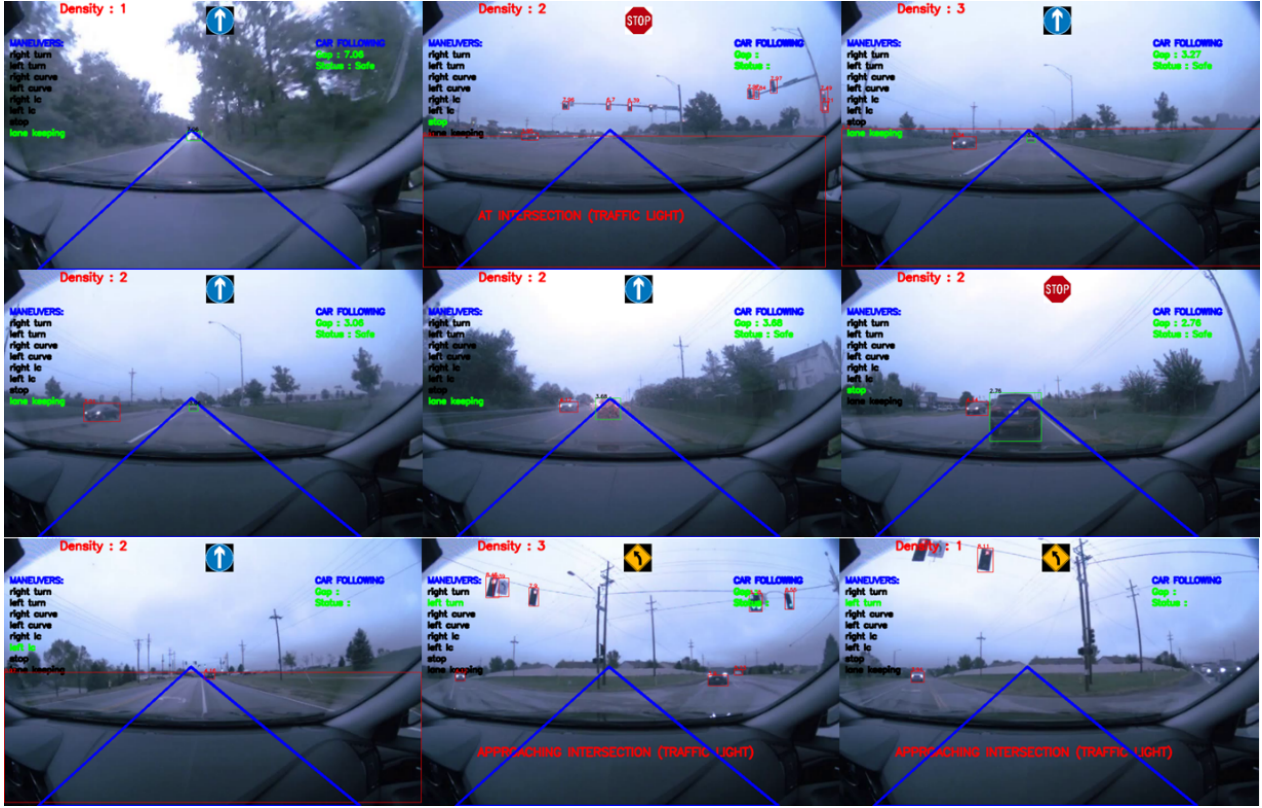

Figure 9: A framework for detecting leading vehicles, safe car-following distance and driver maneuvers.

*5.3 Comparison of Car Following Distances of Different Driving Groups*

As shown in the probability density plot in Figure 10, we can compare the car following distance of the four participants studied in this project. On average, the car following distance can range from greater than zero to an observable value of just under ten meters, with the majority of numbers falling somewhere in the middle. Additionally, it was discovered that the elderly drivers (man and woman) maintained greater car following distances than the two younger drivers. Even though the data used for this analysis does not represent a substantial proportion of each driving group, we can demonstrate from the results that differences between the older drivers and younger drivers can be observed and estimated from our framework, with older drivers being more safety-conscious than younger drivers from our preliminary results. The authors are also cognizant of the fact that this conclusion may or may not hold true if large samples of each driving group are processed and analyzed further but are looking into the potential of this line of analysis for further study. This preliminary analysis is necessary for defining potential disparities between demographic subgroups and accounting for this in the predictive modeling stage of the study.

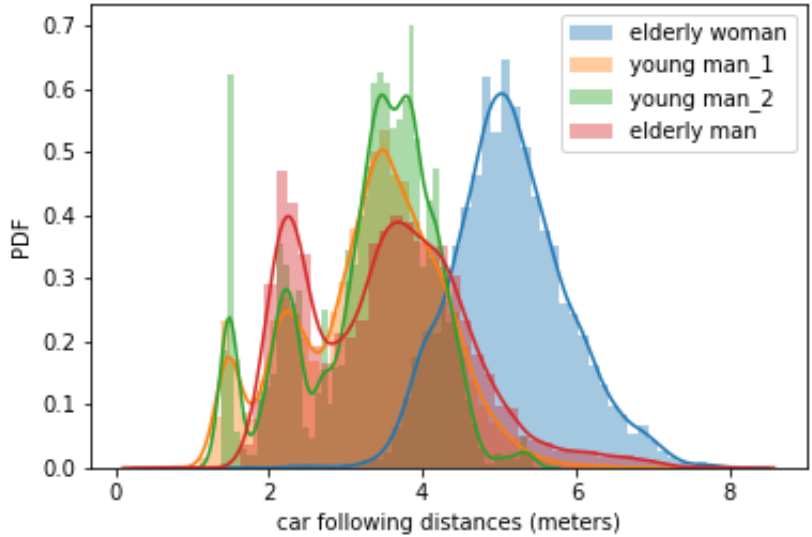

Figure 10: A PDF of car-following distances of different individuals

To further investigate the observations made from Figure 10, statistical t-tests were performed on pairs of driving groups to determine if there were significant differences in the mean car-following distance between the various driving groups. This test is necessary in order to determine if different demographic groups exhibit the same car-following behavior. We assume an equal variance in our analysis if the ratio of variances between any two driving groups is 4:1. The fundamental assumption (null hypothesis) for the test is that the mean car following distance of different driving groups is the same. A significant level ($\alpha$) 0.05 is used to infer if there is enough evidence to reject the null hypothesis. Mathematically we formulated the test as,
Let,

$$\mu_i = the\ mean\ car\ following\ distance\ for\ group\ i$$

$$\alpha = significant\ level$$

then,

$$Null\ Hypothesis:\ H_0 \rightarrow \mu_1 = \mu_2$$

$$Alternate\ Hypothesis:\ H_a \rightarrow \mu_1 \neq \mu_2$$

If the computed p-value < $\alpha$, then we have enough evidence to reject the null hypothesis, else we do not reject the null hypothesis.

With the exception of the comparison between the two young drivers, which yielded a p-value of 0.84, all other comparisons yielded a p-value of 0.00, as shown in Table 2. Given that the p-value for comparing the two young drivers was 0.84, we can conclude that there was insufficient evidence to reject the null hypothesis; thus, the two young drivers' mean car following distance is identical. Also, the results indicate that the elderly female driver and the elderly male driver have different car following behaviors. In addition, by comparing the car following distances of young and elderly drivers, we discovered that the two groups were also different.

Table 4. 2: Statistical T-test Analysis of different individuals

| comparison | P-value | Decision |
| --- | --- | --- |
| Elderly woman vs Elderly man | 0.00 | Reject the null hypothesis |
| Elderly woman vs Young man 1 | 0.00 | Reject the null hypothesis |
| Elderly woman vs Young man 2 | 0.00 | Reject the null hypothesis |
| Elderly man vs Young man 1 | 0.00 | Reject the null hypothesis |
| Elderly man vs Young man 2 | 0.00 | Reject the null hypothesis |
| Young man 1 vs Young man 2 | 0.84 | Do not reject the null hypothesis |
| Young vs Elderly | 0.00 | Reject the null hypothesis |

*5.4 Comparing the Acceleration-Deceleration Behavior of Young Drivers and Elderly Drivers*

As indicated by our earlier analysis, there is a significant disparity between the car-following distances of young and elderly drivers. In this section we explore further the potential to investigate the aggressiveness of each driving group. Figure 11 below presents the relative velocity–acceleration mapping obtained from the two driving groups (young drivers and elderly drivers). The two driving groups have similar data length (young drivers, 1800 sec; elderly drivers, 2000 sec). The deceleration of young drivers (Figure 11a) is greater than that of elderly drivers (Figure 11b) when their vehicle approaches the leading vehicle. This hard deceleration could be used to infer aggressive driving of young drivers. When the distance between vehicles is opening, the acceleration of young drivers is greater than that of elderly drivers. Thus, the acceleration-deceleration rate of young drivers indicates a tendency opposite that of elderly drivers.

Although some disparity in car following behavior between the demographic subgroups is observed, the final modeling phase did not account for a demographic category component with regards to the factors utilized in predicting the acceleration of the leading vehicle. A driver demographic subgroup variable will be included in further studies when data points from a larger sample group of diverse drivers have been extracted and analyzed to further validate the preliminary observed difference in driving behaviors between the subgroups.

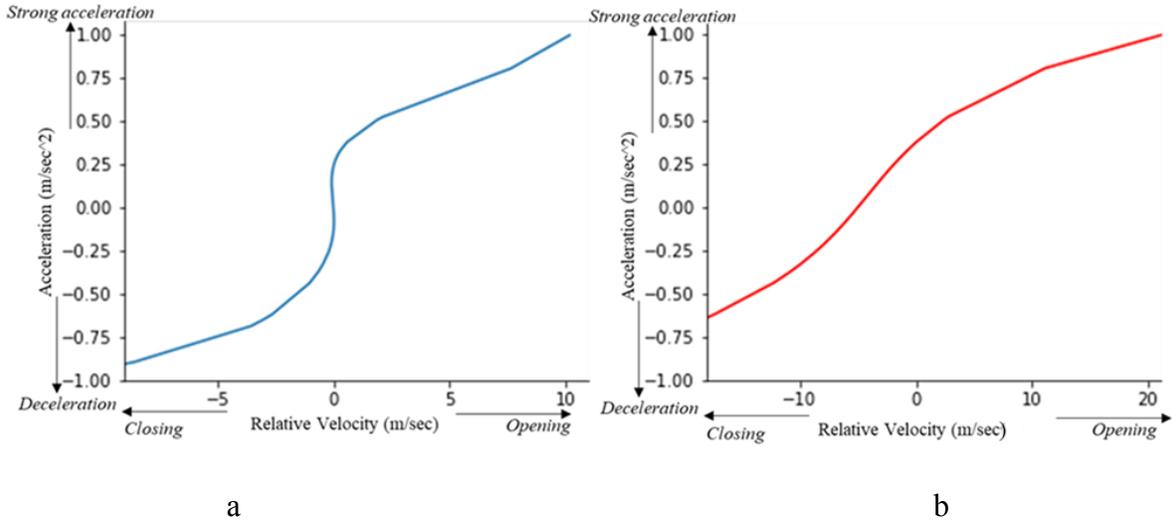

a                                                                 b

Figure 11: Acceleration against relative velocity of two driving groups a) Young drivers b) Elderly drivers

*5.5 Modeling the Acceleration of the Ego-Vehicle*

Modeling the acceleration of the ego-vehicle enables us to comprehend how the ego vehicle accelerates and decelerates depending on its distance from the leading vehicle. In Figure 12, we show a general relationship between acceleration of car-following distance of the ego-vehicle. Generally, deceleration occurs when the ego-vehicle is close to the leading vehicle whereas acceleration occurs when the distance between the ego-vehicle and the leading vehicle is greater.

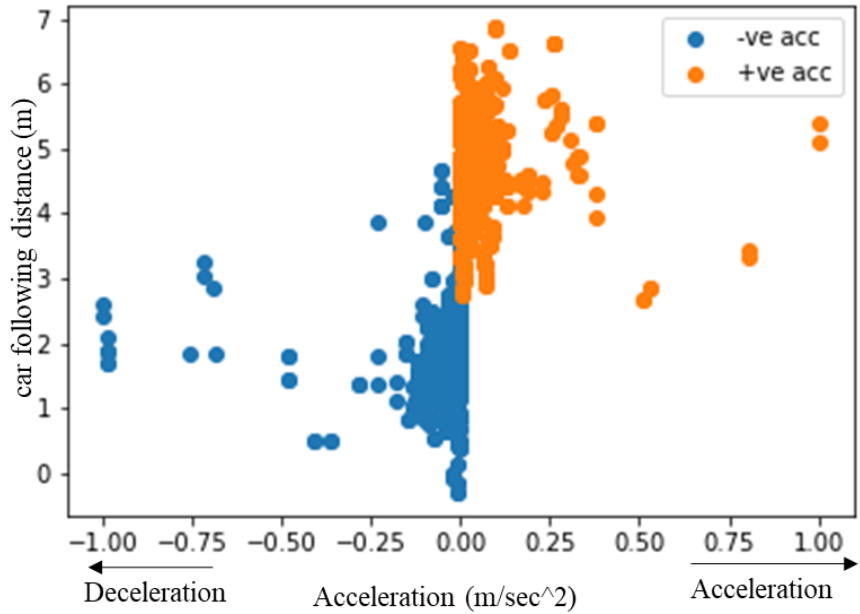

Figure 12: A general relationship between acceleration of car-following distance of the ego-vehicle

To develop the acceleration model for the ego-vehicle, an Extreme Gradient Boosting (XGBoost) algorithm was used. XGBoost is a variant of gradient boosting machines that predicts errors using Gradient Boosting Trees (GBtree). It begins with a simple predictor that predicts an arbitrary number (typically 0.5) irrespective of the input. Typically, the model begins with a high error rate and gradually learns to reduce it. Training an error prediction model in XGBoost does not involve optimizing the predictor on (feature, error) pairs. The data used to train the model was divided into training (5834 data points) and test (2,500 data points) samples in a ratio of 0.7:0.3.

From the results of the model building, relative velocity clearly stands out as the most important predictor of the ego-vehicle acceleration compared to the car following distance as shown in the Figure 1.3.

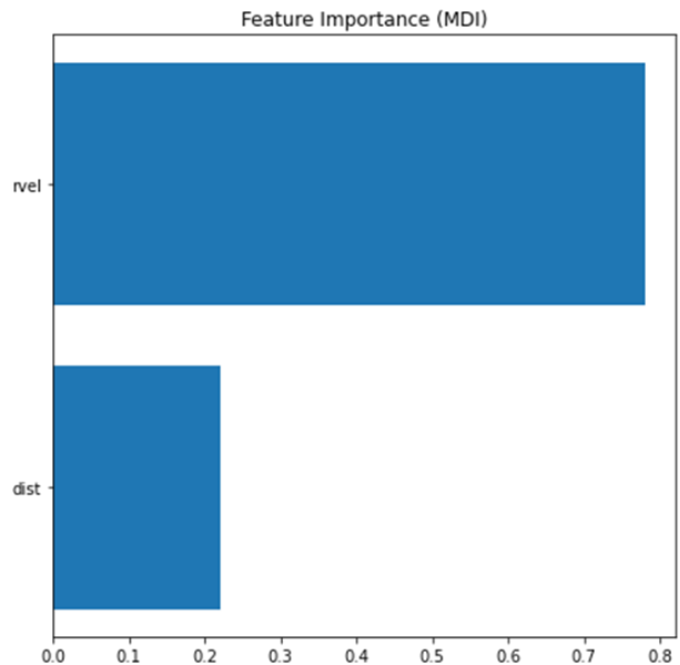

Figure 13: A plot of feature importance from the developed model

Subsequently, the developed model was used to predict the acceleration of the ego-vehicle, as shown in Figure 14. The model was able to detect the general trend of the acceleration of the ego-vehicle as well as the peaks in the trend. The performance of the model was evaluated using the root mean square error (RMSE). The RMSE of the model was 0.0245. A model with RMSE close to zero is commonly regarded as a good model, and since our developed model is in the hundredth decimal, it can be considered to be performing well.

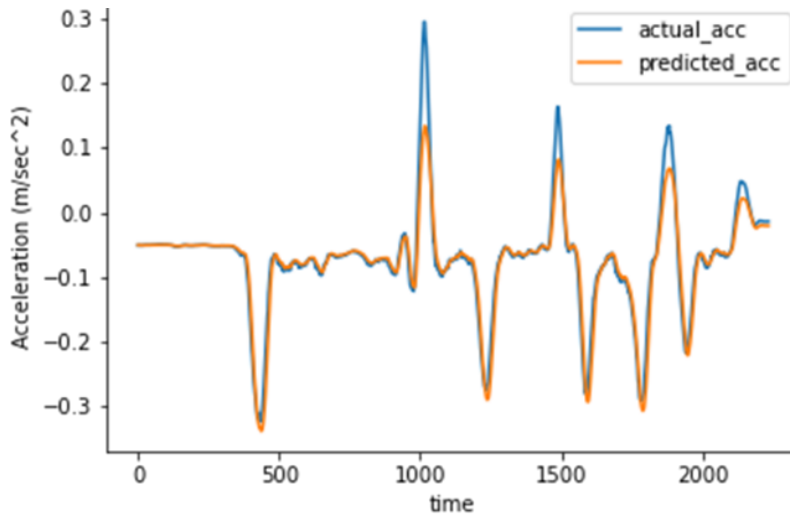

Figure 14: A plot of actual acceleration versus predicted acceleration of the ego-vehicle

*5.6 Modeling the Acceleration of the Leading Vehicle*

The acceleration of the leading vehicle was modeled similar to the acceleration of the ego-vehicle using XGBoost. The difference here is that the acceleration of the ego-vehicle was an input variable to this model in addition to car-following distance and relative velocity. Similar to the model developed above, the data used to train the model was divided into training (5834 data points) and test (2,500 data points) samples in a ratio of 0.7:0.3. The results of the model showing the feature importance are shown in Figure 15. The car-following distance and relative velocity were found to be the most important features to predict the acceleration of the leading vehicle. The acceleration of the ego-vehicle was found not to be influential in predicting the acceleration of the leading vehicle.

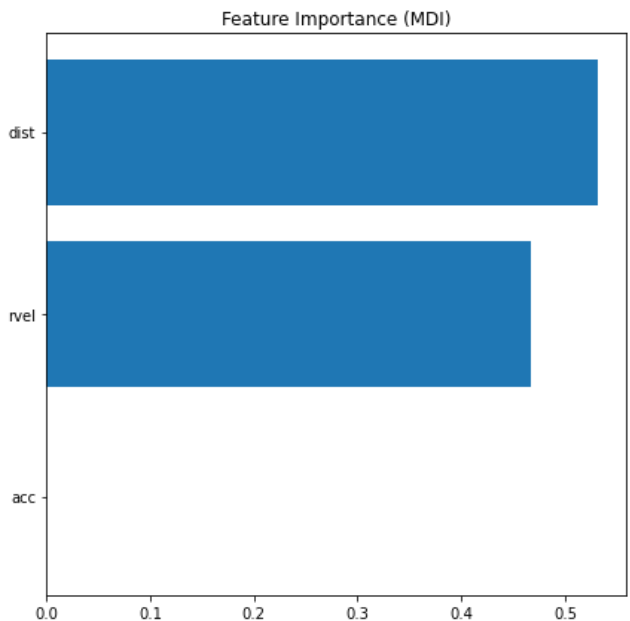

Figure 15: A plot of feature importance from the developed model

As shown in Figure 16, the developed model was used to predict the acceleration of the leading vehicle. The model's generalization ability has been demonstrated. The model was able to detect the general trend of the leading vehicle's acceleration as well as the peaks in the trend. The performance of the model was evaluated using the root mean square error (RMSE). The RMSE of the model was 0.0105. A model with RMSE close to zero is commonly regarded as a good model, and since our developed model is in the hundredth decimal, it can be considered to be performing well.

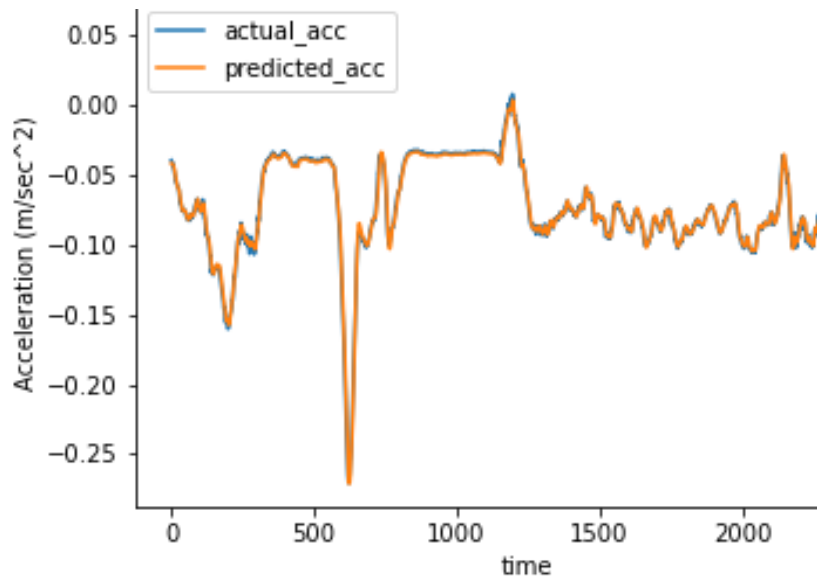

Figure 16: A plot of actual acceleration versus predicted acceleration of the leading-vehicle

## 6. Conclusion

Rear-end crashes are the most common type of accidents encountered on the highway. These crashes have a significant negative impact on the flow of traffic and typically result in serious repercussions. To gain a better understanding of these scenarios from a practical standpoint, it is necessary to accurately model car-following behaviors that lead to rear-ends crashes. Numerous studies have been conducted to model the car-following behaviors of drivers; however, the majority of these studies have relied on simulated data that may not accurately represent real-world incidents. Also, most studies are limited to modeling the acceleration of the ego-vehicle which is insufficient to explain the behavior of the ego-vehicle.

As a result, the current study attempts to address these issues by developing a framework for extracting features pertinent to comprehending the behavior of drivers in naturalistic environments. Moreover, the study modeled the acceleration of both the ego-vehicle and the leading vehicle through the development of an AI framework capable of extracting the parameters from NDS videos required to model the behavior (acceleration) of both vehicles. Additionally, the study performed longitudinal studies of the car-following behavior of different demographics in a naturalistic environment.

The result from the analysis shows that young individuals are more likely to be aggressive drivers compared to elderly drivers. Also, in modeling the acceleration of the ego-vehicle, relative velocity between the ego-vehicle and the leading vehicle was found to be more important than the distance between the two vehicles (car-following distance). This means that the accelerations of drivers in ego vehicles are affected more by the relative velocity between them and the leading vehicle. The opposite occurred when modeling the acceleration of the leading vehicle.

6.1 Limitations and Recommendations

The current study has the following limitations:
The heuristic method utilized by the study to track the leading vehicle is not always effective at intersections or when the driver is negotiating a curve. Future research will employ capitalizing on tracking algorithms to define and follow the changing trajectory of the leading vehicle in such instances.